\pdfoutput=1

\documentclass[11pt,a4paper]{article}
\usepackage[hyperref]{naaclhlt2019}
\usepackage{times}
\usepackage{latexsym}

\def\aclpaperid{1785}

\usepackage{url}

\aclfinalcopy 

\usepackage{comment}
\usepackage{soul}

\usepackage{helvet}  
\usepackage{courier}  
\usepackage{url}  
\usepackage{graphicx}  
\usepackage{booktabs}
\usepackage{tabularx}
\usepackage{comment}

\usepackage{microtype}
\usepackage{graphicx}
\usepackage{subfigure}
\usepackage{booktabs} 
\usepackage{wrapfig}

\usepackage[utf8]{inputenc} 
\usepackage[T1]{fontenc}    
\usepackage{url}            
\usepackage{amsfonts}       
\usepackage{nicefrac}       
\usepackage{subfigure}
\graphicspath{{figures/}}
\usepackage{xcolor}
\usepackage{adjustbox}
\usepackage{bm}
\usepackage{amsmath}
\usepackage{amssymb}
\usepackage{algorithm}
\usepackage{algorithmic}
\usepackage{multirow}
\usepackage{adjustbox}

\usepackage{bm}

\newcommand{\RN}[1]{%
	\textup{\lowercase\expandafter{\it \romannumeral#1}}%
}

\usepackage{algorithm}
\usepackage{algorithmic}

\usepackage{amsmath}
\usepackage{amsfonts}
\usepackage{amssymb}

\newcommand{\beq}{\vspace{0mm}\begin{equation}}
\newcommand{\eeq}{\vspace{0mm}\end{equation}}
\newcommand{\beqs}{\vspace{0mm}\begin{eqnarray}}
\newcommand{\eeqs}{\vspace{0mm}\end{eqnarray}}
\newcommand{\barr}{\begin{array}}
\newcommand{\earr}{\end{array}}

\newcommand{\bv}{{\boldsymbol b}}
\newcommand{\cv}{{\boldsymbol c}}
\newcommand{\dv}{{\boldsymbol d}}
\newcommand{\ev}{{\boldsymbol e}}
\newcommand{\fv}{{\boldsymbol f}}
\newcommand{\Fv}{{\boldsymbol F}}

\newcommand{\hv}{{\boldsymbol h}}
\newcommand{\iv}{{\boldsymbol i}}

\newcommand{\lv}{{\boldsymbol l}}

\newcommand{\ov}{{\boldsymbol o}}

\newcommand{\rv}{{\boldsymbol r}}
\newcommand{\sv}{{\boldsymbol s}}

\newcommand{\vv}{{\boldsymbol v}}

\newcommand{\Wv}{{\boldsymbol W}}

\newcommand{\Uv}{{\boldsymbol U}}

\newcommand{\yv}{{\boldsymbol y}}
\newcommand{\Xv}{{\boldsymbol X}}

\usepackage[bottom]{footmisc}

\title{Figure Captioning with Reasoning and Sequence-Level Training}
\author{Charles Chen$^{1,3}$, Ruiyi Zhang$^2$, Eunyee Koh$^3$, Sungchul Kim$^3$,\\ 
	\textbf{Scott Cohen$^3$, Tong Yu$^4$, Ryan Rossi$^3$, Razvan Bunescu$^1$}  \\
	$^1$Ohio University, $^2$Duke University, $^3$Adobe Research, $^4$Carnegie Mellon University \\
	{\tt lc971015@ohio.edu} \\}

\date{}

\begin{document}
\maketitle

\begin{abstract}
Figures, such as bar charts, pie charts, and line plots, are widely used to convey important information in a concise format. They are usually human-friendly but difficult for computers to process automatically. 
In this work, we investigate the problem of figure captioning where the goal is to automatically generate a natural language description of the figure. 
While natural image captioning has been studied extensively, figure captioning has received relatively little attention and remains a challenging problem.
First, we introduce a new dataset for figure captioning, FigCAP, based on FigureQA. Second, we propose two novel attention mechanisms. To achieve accurate generation of labels in figures, we propose Label Maps Attention. To model the relations between figure labels, we propose Relation Maps Attention.  Third, we use sequence-level training with reinforcement learning in order to directly optimizes evaluation metrics, which alleviates the exposure bias issue and further improves the models in generating long captions. Extensive experiments show that the proposed method outperforms the baselines, thus demonstrating a significant potential for the automatic captioning of vast repositories of figures.

\end{abstract}
\section{Introduction}
Understanding images has been an important area of investigation within computer vision and natural language processing. Recent work has shown excellent performance on a number of tasks, especially image captioning and Visual Question Answering (VQA).
Figures, as a specific type of images, convey useful information, such as trends, proportions and values, in a concise format. People can understand these attributes at a glance. Therefore, people usually use figures (e.g., bar chart, pie chart, and line plot) in documents, reports and talks to efficiently communicate ideas.
%
Figure captioning aims at generating a natural language description of a figure, by inferring potential logical relations between elements in the figure. 
This topic is interesting from the artificial intelligence perspective: the machines would extract the relations between the labels in the figures based on visual intuitions, instead of reconstructing the source data, \emph{i.e.}, inverting the visualization pipeline. 

While natural image captioning has been extensively studied in computer vision, figure captioning has received little attention. Depending on the user case, the generated caption may be a high-level description of the figure, or it may include more details such as relations among the data presented in the figure. 
There are two major challenges in this task. First, it requires an understanding of the labels and relations among labels in a figure. Second, the figure captions typically contains a few sentences, which are usually longer than the captions for natural images (e.g. MSCOCO dataset~\cite{lin2014microsoft}).  As a long-text-generation task, figure captioning will accumulate more errors as more words predicted.


A similar problem of understanding figures is 
VQA. However, figure captioning distinguishes itself from VQA in two important aspects. First, the input is different. The input to a VQA system consists of an image/figure to be queried and a question. A figure captioning system automatically generates a description of the figure, which can be regarded as a self-asking VQA task. Second, the output of a VQA system is the answer to the given question, commonly containing only a few words. In contrast, a figure captioning system usually produces a few sentences. 

In this paper, we investigate the problem of figure captioning. 
Our main contributions in this work are:
\begin{itemize}
    \item We introduce a new dataset for figure captioning called FigCAP.
    \item We propose two novel attention mechanisms to improve the decoder's performance. The Label Maps Attention enables the decoder to focus on specific labels. The Relation Maps Attention is proposed to discover the relations between figure labels. 
    \item We utilize sequence-level training with reinforcement learning to handle long sequence generation and alleviate the issue of exposure bias.
    \item Empirical experiments show that the proposed models can effectively generate captions for figures under several metrics. 
\end{itemize}






\section{Related Work}
\paragraph{Image Captioning} The existing approaches for image captioning largely fall into two categories: top-down and bottom-up. The bottom-up approaches first output key words describing different aspects of an image such as visual concepts, objects, attributes, and then combines them to sentences.
~\cite{farhadi2010every,kulkarni2011baby,elliott2013image,lebret2014simple,fang2015captions} lie in this category. The successful application of deep learning in natural language processing, for example, machine translation, motivates the exploration on top-down methods, such as ~\cite{mao2014deep,donahue2015long,jia2015guiding,vinyals2015show,xu2015show}. These approaches formulate the image captioning as a machine translation problem, directly translating an image to sentences by utilizing the encoder-decoder framework.
The approaches based on deep neural networks proposed recently largely fall into this category.

\paragraph{Visual Question Answering} Another task related to the figure captioning problem is VQA \cite{kafle2017visual}, which is to answer queries in natural language about an image. The traditional approaches \cite{antol2015vqa,gao2015you,kafle2016answer,zhou2015simple,saito2017dualnet} train a linear classifier or neural network with the combined features from images and questions. Bilinear pooling or related techniques are further proposed to efficiently and expressively combine the image and question features \cite{fukui2016multimodal,kim2016multimodal}. Spatial attention was used to adaptively modify the visual features or local features in VQA \cite{xu2016ask,yang2016stacked,ilievski2016focused}. Bayesian models were used to discover the relationships between the features of the images, questions and answers \cite{malinowski2014multi,kafle2016answer}.
Previous works \cite{Andreas_2016_CVPR,andreas2016learning} also decompose VQA into several sub-problems and solve these sub-problems individually.


%
\paragraph{Figure VQA}
VQA has been used to answer queries in natural language about figures. 
Kahou \emph{et al.} \cite{kahou2017figureqa} introduced FigureQA, a novel visual reasoning corpus for VQA task on figures. On this dataset, relation network \cite{santoro2017simple} has strong performance among several models. Kafle \emph{et al.} \cite{kafle2018dvqa} presented DVQA, a dataset used to evaluate bar chart understanding by VQA.
On this dataset, multi-output model and SAN with dynamic encoding model have been shown to achieve better performances.

\begin{figure}[htbp]
\centerline{\includegraphics[width=0.43\textwidth]{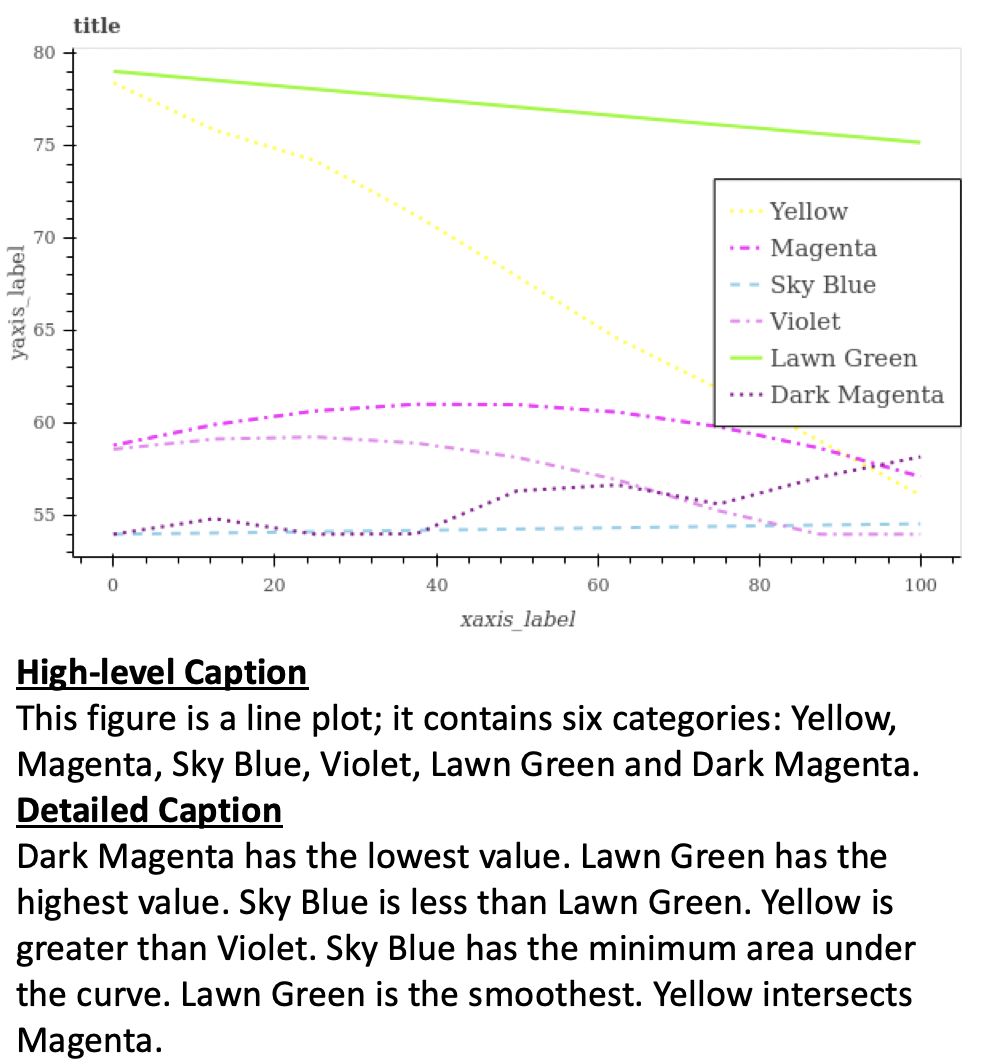}}
\caption{An example in FigCAP. We generate both high-level and detailed captions for the figure.}
\label{fig:prodef}
\end{figure}

\begin{figure*}[htbp]
\centerline{\includegraphics[width=0.85\textwidth]{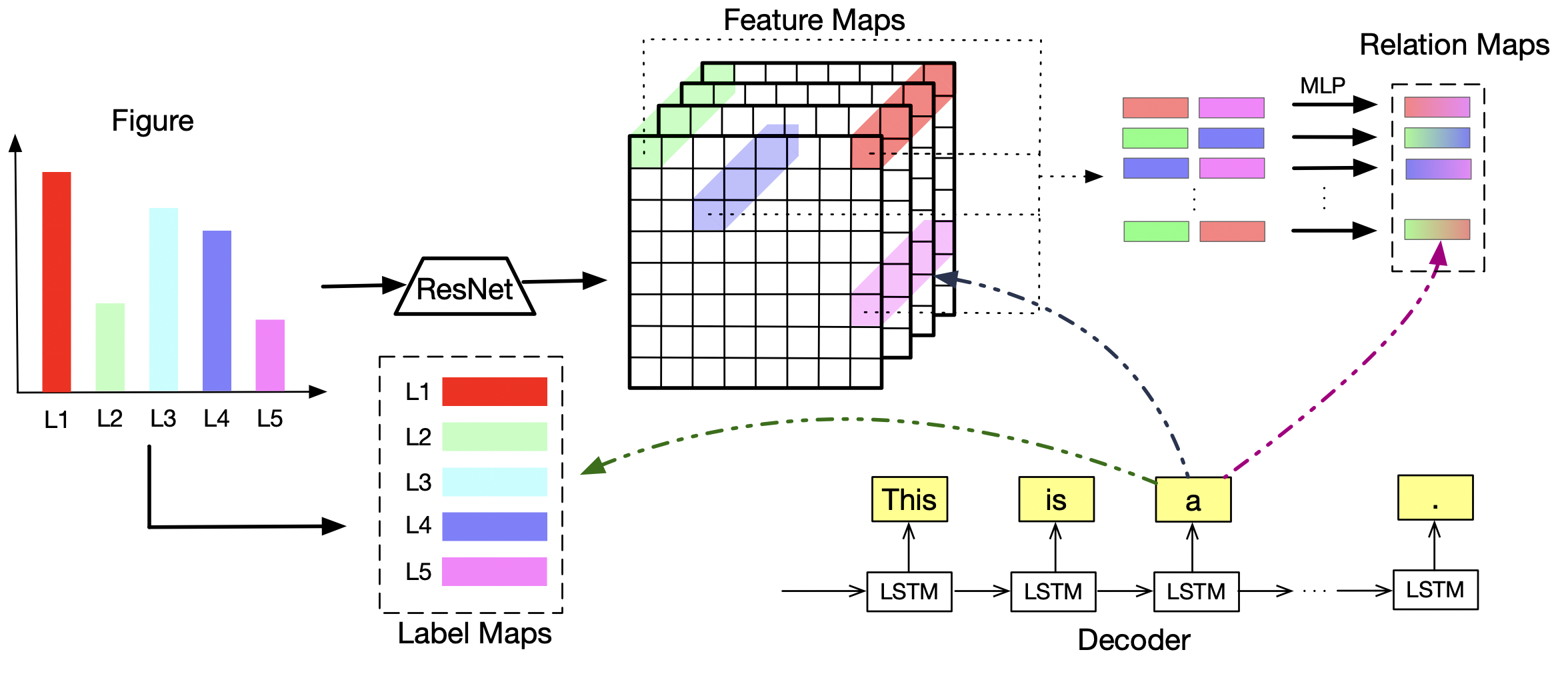}}
\caption{Model overview: Our model takes a figure image as input, encodes it with \textit{ResNet}. Decoder is a LSTM with Attention Models \textit{Att\_F}, \textit{Att\_R} and \textit{Att\_L}. Solid arrow lines show data flows, and dash arrow lines show the attentions. 
}
\vspace{-4mm}
\label{fig:modeloverview}
\end{figure*}

\section{Background}
\subsection{Sequence-Generation Model}
A sequence-generation model generates a sequence $Y=(y_1, \ldots, y_T)$ conditioned on an object $X$, where $y_t\in\mathcal{A}$ is a generated token at time $t$ and $\mathcal{A}$ is the set of output tokens. The length of an output sequence is denoted as $T$, and $Y_{1,\dots,t}$ indicates a subsequence $(y_1, \ldots, y_t)$. The data are given with $(X,Y)$ as pairs to train a sequence-generation model. We denote the output a sequence-generation model as $\hat{Y}$. 

Starting from the initial hidden state $\sv_0$, a RNN produces a sequence of states $(\sv_1, \sv_2,\ldots, \sv_T)$, given a sequence-feature representation $(e(y_1), e(y_2), \dots, e(y_T))$, where $e(\cdot)$ denotes a function mapping a token to its feature representation. Let $\ev_t \triangleq e(y_t)$. The states are generated by applying a transition function $h: \sv_t=h(\sv_{t-1}, \ev_t)$ for $T$ times. The transition function $h$ is implemented by a cell of an RNN, with popular choices being the Long Short-Term Memory (LSTM~\cite{hochreiter1997long}) and Gated Recurrent Units~(GRU~\cite{cho2014learning}). We use LSTM in this work. To generate a token $\hat{y}_t\in\mathcal{A}$, a stochastic output layer is applied on the current state $\sv_t$: 
\begin{align}\nonumber
\hat{y}_t&\sim\text{Multi}(1, \text{softmax}(g(\sv_{t-1}))),\\
\nonumber
\sv_t &= h(\sv_{t - 1}, e(\hat{y}_{t}))\,
\end{align}   

where $\text{Multi}(1,\cdot)$ denotes one draw from a multinomial distribution, and $g(\cdot)$ represents a linear transformation.
Since the generated sequence $Y$ is conditioned on $X$,  one can simply start with an initial state encoded from $X$: $\sv_0=\sv_0(X)$~\cite{bahdanau2016actor,cho2014learning}. 
Finally, a conditional RNN can be trained for sequence generation with gradient ascent by maximizing the log-likelihood of a generative model.
\subsection{Sequence-Level Training}
Sequence-generation models are typically trained with “Teacher-Forcing”, which maximizes the likelihood estimation (MLE) of the next ground-truth word given the previous ground-truth word. This approach accelerates the convergence of training, but introduces exposure bias~\cite{ranzato2015sequence}, caused by the mismatch between training and testing. The error will accumulate during testing, and this problem becomes more severe when the sequence become longer. 

Sequence generation with reinforcement learning (RL) can alleviate exposure bias and improve the performance by directly optimizing the evaluation metrics via sequence-level training. Instead of training in word-level as MLE, sequence-level training is guided by the reward of the sequence. Variants of this method include adding actor-critic~\cite{bahdanau2016actor} or  self-critical baselines~\cite{rennie2016self,anderson2017bottom} to stabilize the training. Besides,~\cite{luo2018discriminability} used image retrieval model to discriminate the generated and reference captions combined with sequence-level training.

\section{Problem Definition and Dataset} \label{l:dataset}

\paragraph{Figure Captioning} This task aims at producing descriptions with essential information for a given figure. The input is a figure and the expected output is the caption for this figure. The caption may contains high-level information only, such as figure type, number of labels, and label names. This is to give the users a rough idea of the content in the figure. Or the caption may contain more details, such as the relations among labels (\emph{e.g.}, A is larger than B, C has the maximum area under the curve). This is to give the users a deep understanding of the logic demonstrated in the figure. Depending on the use cases, the tasks of figure captioning can be categorized into (i) generating high-level captions for figures and (ii) generating detailed captions for figures.

\paragraph{FigCAP} There are some public datasets from previous work on figure understanding, such as FigureSeer~\cite{siegel2016figureseer}, DVQA\cite{kafle2018dvqa} and FigureQA~\cite{kahou2017figureqa}. FigureSeer contains figures from research papers, while plots in both DVQA and FigureQA are synthetic. Due to the synthetic nature, one can generate as many figures, accompanied by questions and answers as he wants. In this sense, the size of FigureSeer is relatively small compared to DVQA and FigureQA, though its figures come from real data. In terms of figure type, FigureQA contains vertical and horizontal bar charts, pie charts, line plots, and dot-line plots while DVQA has only bar charts. Also, reasoning ability is important for captioning approaches to generate good quality captions. Note that FigureQA is designed for visual reasoning task. Considering the above factors, we generate our dataset FigCAP based on FigureQA. 

FigCAP consist of figure-caption pairs where figures can be generated by the method introduced in~\cite{kahou2017figureqa} and captions are based on corresponding fundamental data, \emph{i.e.}, they are ground truth captions (reference captions). 
Note that a human would obviously not describe a figure with exactly the same sentences. 
To increase diversity of reference captions, we design templates to paraphrase sentences. Table \ref{tab:example} lists selected templates we use to paraphrase sentences.

\begin{table*}[ht]
\begin{center}
\begin{tabular}{lcccccc}
\toprule[1.2pt]
This figure includes N labels: A, B, C..; A is the maximum...\\
There are N labels in this TYPE; their names are A, B, C..\\
This is TYPE; it has N labels: A, B, C, D...; A is larger than B, B is the maximum...\\
This figure is TPYE; it contains N categories; their names are A, B, C, D...; A is larger than B\\
There are N different labels in this line plot, with labels A, B...; D has the largest area under the curve...\\
This figure is TYPE; there are N categories in it; their names are A, B, C...; C is the minimum...\\
There are N different bars in this TYPE: A, B, C, D...; C is the minimum...\\
It is a dot line plot, with N lines: A, B, C, D...; C is the minimum...\\
This figure is a dot line plot; there are N lines; their names are A, B, C...; C is the minimum...\\
There are N categories in this dot line plot: A, B, C...\\
\bottomrule[1.2pt]
\end{tabular}
\caption{The selected templates for generation captions from QA dataset.}
\label{tab:example}
\end{center}
\end{table*}

With these templates, we develop two datasets, \textbf{FigCAP-H} and \textbf{FigCAP-D} for two different use cases. FigCAP-H contains \underline{\textbf{H}}igh-level descriptions for figure captions. In contrast, FigCAP-D contains \underline{\textbf{D}}etailed descriptions for figure captions. Both FigCAP-H and FigCAP-D have five different types of figures: horizontal bar chart, vertical bar chart, pie chart, line plot and dotted line plot. The numbers of each type of figures are roughly the same for each of them. Table~\ref{tab:statistics} shows the numbers of figure-caption pairs for both datasets. Their sizes are similar to the setting in (Gan et al., 2017). Note that since both figures and captions are synthetic, the figure-caption pairs can be generated as many as needed. 
\begin{table}[ht]
\renewcommand{\arraystretch}{1.1} 
\begin{center}
\begin{tabular}{lccccccc}
\toprule[1.2pt]
Datasets & Training & Validation & Testing \\
\midrule
FigCAP-H & 99,360 & 5,000 & 5,152 \\
FigCAP-D & 99,360 & 5,000 & 5,152 \\
\bottomrule[1.2pt]
\end{tabular}
\caption{Statistics for FigCAP-H and FigCAP-D.}
\label{tab:statistics}
\end{center}
\end{table}
An example of captions for Figure~\ref{fig:prodef} is the following. 

\textit{``\ul{This is a line plot. It contains 6 categories. Dark Magenta has the lowest value. Lawn Green has the highest value. 
}[Sky Blue is less than Lawn Green. Yellow is greater than Violet. Sky Blue has the minimum area under the curve. Lawn Green is the smoothest. Yellow intersects Magenta.]``}

The words underlined are high level captions of the figure. The words in square brackets are detailed captions of the figure, which describes the relationships among the labels of categories represented by plotted lines.

Figure captioning using the FigCAP data is more challenging than natural image captioning for two main reasons.
First, the sentences in figure captioning are much longer, compared with natural image captioning.
Second, the logical information is much more important and complex, yet is very difficult and challenging to extract from figures.
Another important and challenging problem is how to capture the key information and insights from the figure automatically, e.g., humans can derive key insights from the figure by making inferences based on the logical and semantic information in the figure.

\section{The Proposed Models}

We describe the proposed model for figure captioning, as illustrated in Figure~\ref{fig:modeloverview} . The model generally follows an encoder-decoder structure. The encoder is a Residual Network~\cite{he2016deep} which extracts feature maps from the given figures. Reasoning network, built upon the feature maps, produces relation maps which embed logical information in the given figure. We use LSTM~\cite{hochreiter1997long} for decoding. With our proposed attention models, the decoder may optionally attend to the label maps, feature maps and/or relation maps. The objective of figure captioning is to maximize likelihood or total rewards. The details of each component will be presented in the following subsections.

\subsection{Captioning Model}

Similar to the approaches in ~\cite{rennie2016self,karpathy2015deep}, we use the following neural networks for figure captioning. The figure $\Xv$ is used as the input of a \textit{ResNet}. 
\begin{align*}
\Fv = ResNet({\Xv})
\end{align*}

The output of the \textit{ResNet} (Feature Maps) $\Fv$ is used to initialize a LSTM: 
\begin{align*}
\cv_{0} = \sigma(\Wv_{Ic} \Fv) \\
\hv_{0} = \sigma(\Wv_{Ih} \Fv)
\end{align*}
$\sigma(.)$ is the sigmoid function. The caption is preprocessed with a BOS token in the beginning and a EOS token in the end. We use the one-hot vector $\bm{1_{y,t}}$ to represent the word $y_t$, and the encoding $\bm{1_{y,t}}$ is further embedded by a linear embedding $\bm{E}$. 
\begin{align*}
\ev_{t} & = e(y_t) = \bm{E 1_{y,t}}, t > 0\\
\ev_{0} & = \bm{0}, otherwise
\end{align*}


The word vector $\ev_t$ and context vector $\dv_t$ (See Section: Attention Models for Figure Captioning) are used as the input of the LSTM.
The signals for input gate, forget gate and output gate are
\begin{align*}
  \iv_t &= \sigma(\Wv_{iy} \ev_t + \Wv_{ih} \hv_{t-1} + \Wv_{id} \dv_{t} + \bv_i) \\
  \fv_t &= \sigma(\Wv_{fy} \ev_t + \Wv_{fh} \hv_{t-1} + \Wv_{fd} \dv_{t} + \bv_f) \\
  \ov_t &= \sigma(\Wv_{oy} \ev_t + \Wv_{oh} \hv_{t-1} + \Wv_{od} \dv_{t} + \bv_o),
\end{align*}
respectively. $\dv_t$ is the context vector, $\sigma(.)$ is the sigmoid function, and $\hv_{t-1}$ is the output of hidden layer in the LSTM. With the signals for input gate, forget gate and output gate, $\hv_{t}$ is computed as:
\begin{align*}
  \cv_t &= \iv_t\odot\phi(\Wv_{cy}^{\otimes} \ev_t + \Wv_{ch}^{\otimes} \hv_{t-1} + \Wv_{cd}^{\otimes} \dv_t + \bv_c^{\otimes} ) \\
     & + \fv_t\odot \cv_{t-1}\\
  \hv_t &= \ov_t\odot \text{tanh}(\cv_t)
\end{align*}
where $\dv_t$ is the context vector, $\text{tanh}(.)$ is the hyperbolic tangent function, and $\phi(.)$ is the maxout non-linearity.

We use both the context vector $\dv_t$ and $\hv_t$ to predict the next word $y_t$:
\begin{align*}
\tilde{\yv}_t &= \sigma(\Wv_h \hv_t + \Wv_d \dv_t) \\
y_t & \sim \text{softmax}(\tilde{\yv}_t)
\end{align*}
We illustrate details for computing context vector $d_t$ with multiple attention mechanism in next section.







\subsection{Attention Models for Figure Captioning}
Attention mechanism has been widely used in the encoder-decoder structure to improve the decoding performance. We propose two attention models: Relation Maps Attention (\textit{Att\_R}), and Label Maps Attention (\textit{Att\_L}). We also introduce Feature Maps Attention (\textit{Att\_F}). Context vector $\dv_t$ can be computed from one of them, or combination of them. 

\subsubsection{Feature Maps Attention \textit{Att\_F}}
Feature Maps Attention Model takes Feature Maps $\Fv$ ($\Fv$ contains $m$ feature vectors; $\Fv \in R^{m\times d}$) and the hidden state $\hv_{t-1}$ of LSTM as input. For each feature $\fv_j$ in $\Fv$, it computes a score between $\fv_j$ and $\hv_{t-1}$. With these scores as weights, it computes the context vector $\cv_t$
as the weighted sum of all features in the feature maps. Equation~\ref{eq:attention1} defines Feature Maps Attention Model:
\begin{align}
e_{tj} & = Att\_F(\hv_{t-1}, \fv_j) \label{eq:attention1}  \\
 & = \vv^T_a \tanh(\Wv_a \fv_{j} + \Uv_a \hv_{t-1}) \nonumber  \\
\alpha_{tj} & = \frac{\exp(e_{tj})}{\sum_{k=1}^{m}\exp(e_{tk})},
\:\:\:\:\:\cv_t = \sum_{j=1}^{m}\alpha_{tj} \cdot \fv_j \nonumber 
\end{align}
where $\fv_j$ is the $j$-th feature in the feature maps $\Fv$, $\cv_t$ is the context vector and $\alpha_{tj}$ is an attention weight.


\subsubsection{Relation Maps Attention \textit{Att\_R}} 



In order to generate correct captions describing relations among the labels (e.g. A is the maximum, B is greater than C, C is less than D.), it is essential to perform reasoning among labels in a given figure. Inspired by Relation Networks~\cite{santoro2017simple}, we propose the Relation Maps Attention Model (\textit{Att\_R}). We consider each feature vector $\fv_j\in R^{d}$ in the feature maps $\Fv$ as an object. For any two ``objects'', for example, $\fv_i$ and $\fv_j$, we concatenate them and feed the vector into a MLP, resulting in a relation vector $\rv_{ij} \in R^{\hat{d}}$:
\begin{align}
\rv_{ij} & = MLP(concat(\fv_i, \fv_j)), \rv_{ij} \in R^{\Hat{d}}
\end{align}
Therefore, the relation maps $\bm{R}$ contains $m^2$ relation vectors ($m$ is the number of feature vectors in feature maps $\Fv$).   
Given the relation maps $\bm{R}$, at decoding step $t$, \textit{Att\_R} computes the relation context vector $\hat{\cv}_t$ as follows:
\begin{align}
\hat{e}_{tk} & = Att\_R(\hv_{t-1}, \rv_k) \label{eq:attention2}  \\
 & = \vv^T_b \tanh(\Wv_b \rv_{k} + \Uv_b \hv_{t-1}) \nonumber  \\
\beta_{tk} & = \frac{\exp(\hat{e}_{tk})}{\sum_{l=1}^{m^2}\exp(\hat{e}_{tl})} \nonumber,
\:\:\:\:\:\hat{\cv}_t = \sum_{k=1}^{m^2}\beta_{tk} \cdot \rv_k \nonumber 
\end{align}
where $\rv_k$ is the $k$-th relation vector in relation maps $\bm{R}$ and $\beta_{tk}$ is an attention weight.

Note that more complex relationships can be induced from pairwise relations, e.g. A > B and B > C lead to A > C. The relation map $\bm{R}$ obtained from Reasoning Net represents abstract objects that implicitly represent object(s) in the figure, not explicitly represent one specific object like a bar or a line.

\subsubsection{Label Maps Attention \textit{Att\_L}}

We propose Label Map Attention Model (\textit{Att\_L}) where the LSTM attends to Label Map $\bm{L}$ for decoding. Label Map $\bm{L}$ is composed of embeddings of those labels appearing in the figure. If $n$ is the number of labels in the figure, then $\bm{L}$ contains $n$ vectors. 
Let $\lv_j$ be the $j$-th vector in the label maps $\bm{L}$, we define \textit{Att\_L} as follows:
\begin{align}
\tilde{e}_{tj} & = Att\_L(\hv_{t-1}, \lv_j) \label{eq:attention3}  \\
 & = \vv^T_c \tanh(\Wv_c \lv_{j} + \Uv_c \hv_{t-1}) \nonumber,\\
\gamma_{tj} & = \frac{\exp(\tilde{e}_{tj})}{\sum_{j=1}^{n}\exp(\tilde{e}_{tj})} \nonumber,
\:\:\:\:\:\tilde{\cv}_t = \sum_{j=1}^{n}\gamma_{tj} \cdot \lv_j \nonumber 
\end{align}
where $\tilde{c}_t$ is the context vector at time step $t$.

Note that figure labels are also used as inputs. For example, in Figure~\ref{fig:prodef}, $n$ is 6; Yellow, Magenta, Sky Blue, Violet, Lawn Green and Dark Magenta are extracted from it using state-of-the-art computer vision techniques such as Optical Character Recognition (OCR).
Since labels appear in the caption of the input figure, instead to define a new set of vectors to represent the labels in the Label Maps $\bm{L}$, we use a subset of the word embeddings $\bm{E}$. In Figure~\ref{fig:prodef}, embeddings $\ev$ for Yellow, Magenta, Sky Blue, Violet, Lawn Green and Dark Magenta compose its Label Map $\bm{L}$. 



\subsubsection{Context Vector $\dv_t$}
In the captioning model, the decoder can use any combination of \textit{Att\_F}, \textit{Att\_R} and \textit{Att\_L}, or it can use only one of them. For example, if we incorporate all three Attention Models (Eq.\ref{eq:attention1},\ref{eq:attention2},\ref{eq:attention3}) in the caption generation model, the final context vector $\dv_t$, used as input to the decoder, is as follows:
\begin{align}
\dv_t = \text{concat}(\cv_t, \hat{\cv}_t, \tilde{\cv}_t)
\end{align}
We explore different combinations of Attention Models for generating captions. More details are in Experimental Evaluations (Section~\ref{sec:experiment}).

\subsection{Hybrid Training Objective}
\label{sec:loss}

In the traditional method \cite{williams1989learning}, ``Teacher forcing''  is widely used for the supervised training of decoders. Given an object X, it maximizes the likelihood of the target word $y_t$, given the previous target sequences $Y_{t-1}$:
\begin{align}
L_{sl} = -\sum_{t=1}^{T} \log p(y_t | Y_{t-1}, x).
\label{eq:ml-loss}
\end{align}

Due to the exposure bias and indirectly optimizing the evaluation metric, supervised training usually can not provide best results. Besides, the word-level training is difficult to handle the generation with different but reasonable word-orders. As a long-text-generation task, figure captioning will accumulate more errors as more words predicted and diversity will be undermined. 

Sequence-level training with RL can effectively alleviate the mentioned problems, by directly optimizing the sequence-level evaluation metric. We use the self-critical policy gradient training algorithm in our model. Specifically, a sequence $\hat{Y}^b$ is generated by greedy word search, \emph{i.e.}, selecting the word with the highest probability. Then, another sequence $\hat{Y}^s$ is generated by sampling next word $\hat{y}^s_t$ according to the probability distribution of $p(\hat{y}^s_t|\hat{Y}^s_{t-1})$. 
The sampled sequence $\hat{Y}^s$ is an exploration of the policy for generating the caption, and the sequence $\hat{Y}^b$ obtained from greedy search is the baseline. We use CIDEr as the sequence-level evaluation metric and compute CIDEr for $\hat{Y}^s$ and $\hat{Y}^b$, respectively. The reward is defined as the difference of CIDEr between the sampled sequence $\hat{Y}^s$ and greedy sequence $\hat{Y}^b$. Let $r(Y)$ be the CIDEr of sequence $Y$. We minimize the sequence-level loss (i.e. maximizing the rewards):
\begin{equation}
L_{rl} = -(r(\hat{Y}^s) - r(\hat{Y}^b)) \sum_{t=1}^{T} \log p(\hat{y}^s_t | \hat{Y}^s_{t-1}, x)
\label{eq:rl-loss}
\end{equation}
Our model is pretrained with MLE loss to provide more efficient policy exploration. Good explorations are encouraged while poor explorations are discouraged in future generation. However, we found that purely optimizing sequence-level evaluation metric, such as CIDEr, may lead to overfitting. To tackle this issue, we use hybrid training objective in our model, considering both word-level loss $L_{sl}$ provided by MLE (Eq.\ref{eq:ml-loss}) and sequence-level loss $L_{rl}$ computed by RL (Eq.\ref{eq:rl-loss}):
\begin{align}
L_{hybrid} = \lambda L_{rl} + (1 - \lambda) L_{sl},
\label{eq:mixed-loss}
\end{align}
where $\lambda$ is a scaling factor balancing the weights between $L_{rl}$ and $L_{sl}$. In practice, $\lambda$ starts from 1 and slowly decays to 0, then only reinforcement learning loss is used to improve our generator.




\section{Experimental Evaluations}
\label{sec:experiment}
In this section, we validate our proposed models on the FigCAP-H and FigCAP-D. Specifically, we evaluate the models in two use cases: generating high-level captions and generating detailed captions for figures, respectively. We perform an ablation study on the improvements brought by each part of our proposed method. 

\begin{table*}[ht]
\small
\centering
\renewcommand{\arraystretch}{1.1} 
\begin{center}
\begin{tabular}{lccccccc}
\toprule[1.2pt]
&\multicolumn{7}{c}{\textbf{Evaluation Metrics}} \\
{\bf Models} & CIDEr & BLEU1 & BLEU2 & BLEU3 & BLEU4 & METEOR & ROUGE \\
\midrule
CNN-LSTM & 0.232 & 0.332 & 0.255 & 0.201 & 0.157 & 0.188 & 0.270\\
CNN-LSTM+\textit{Att\_F} & 0.559 & 0.333 & 0.262 & 0.210 & 0.168 & 0.209 & 0.334 \\
CNN-LSTM+\textit{Att\_F}+\textit{Att\_L} & {\bf 1.018} &	{\bf0.337} &	{\bf0.269} &	{\bf0.215} &	{\bf0.170} &	{\bf0.227} & {\bf0.368}\\

\bottomrule[1.2pt]
\end{tabular}
\caption{Results for FigCAP-H: High-level Caption Generation.}
\label{tab:res1}
\end{center}
\end{table*}

\begin{table*}[ht]
\small
\renewcommand{\arraystretch}{1.1} 
\begin{center}
\begin{tabular}{lccccccc}
\toprule[1.2pt]
&\multicolumn{7}{c}{\textbf{Evaluation Metrics}} \\
{\bf Models} & CIDEr & BLEU1 & BLEU2 & BLEU3 & BLEU4 & METEOR & ROUGE \\
\midrule
CNN-LSTM & 0.158 & 0.055 & 0.050 & 0.044 & 0.038 & 0.115 & 0.244\\
CNN-LSTM+\textit{Att\_F} & 0.868 & 0.215 & 0.200 & 0.181 & 0.159 & 0.200 &0.401\\
CNN-LSTM+\textit{Att\_F}+\textit{Att\_L} & 0.917 & 0.232 & 0.214 & 0.194 & 0.170 & 0.207 & 0.413 \\
CNN-LSTM+\textit{Att\_All} & 1.036 & 0.312 & 0.290 & 0.264 & 0.233 & 0.231 & 0.468 \\
CNN-LSTM+\textit{Att\_All}+RL & \textbf{1.179} & \textbf{0.404} & \textbf{0.367} & \textbf{0.324} & \textbf{0.270} & \textbf{0.263} & \textbf{0.489} \\
\bottomrule[1.2pt]
\end{tabular}
\caption{Results for FigCAP-D: Detailed Caption Generation. \textit{Att\_All=Att\_F+Att\_L+Att\_R}.}
\label{tab:res2}
\end{center}
\end{table*}

\subsection{Experimental Settings}

We implement the following models with TensorFlow, and conduct experiments on a single nVidia Tesla V100 GPU. For any of them, \textit{ResNet-50} pretrained on ImageNet~\cite{deng2009imagenet} is used as the encoder and a 256-unit LSTM is the decoder.
\begin{itemize}
\item \textbf{CNN-LSTM}: This baseline model uses basic CNN-LSTM structure, without any Attention Model. 
\item \textbf{CNN-LSTM+\textit{Att\_F}}: This model uses \textit{Att\_F} for decoding. Similar model is used in natural image captioning~\cite{xu2015show}.
\item \textbf{CNN-LSTM+\textit{Att\_F+Att\_L}}: This model uses both \textit{Att\_F} and \textit{Att\_L} for decoding.
\item \textbf{CNN-LSTM+\textit{Att\_F+Att\_L+Att\_R}}: This model uses \textit{Att\_F}, \textit{Att\_L} and \textit{Att\_R} for decoding. 
\item \textbf{CNN-LSTM+\textit{Att\_F+Att\_L+Att\_R}+RL}: 
The loss function of this model is described in Section~\ref{sec:loss}. Training with RL can improve the model's performance when handling long captions, which is suitable for FigCAP-D.
\end{itemize}
\smallskip\noindent
All of them are optimized with Adam~\cite{kingma2014adam} on the training set and evaluated on the testing set. We tune hyperparameters on the validation set. Table~\ref{tab:statistics} shows the statistics of our datasets FigCAP-H and FigCAP-D. Appendix A contains more details on experimental settings. Following~\cite{xu2015show} and ~\cite{rennie2016self}, we use CIDEr~\cite{vedantam2015cider}, BLEU1-4~\cite{papineni2002bleu}, METEOR~\cite{banerjee2005meteor} and ROUGEL~\cite{lin2004rouge} as evaluation metrics. Note that we only evaluate models containing \textit{Att\_R} on FigCAP-D since only long captions contain relation information.




\subsection{Results of Generating High-Level Captions}
We evaluate the proposed models for the task of generating high-level captions. 
Compared to generating the detailed captions, generating high-level descriptions is relatively easier. 
We do not need to model the relations between the labels in the figures. Besides, the high-level captions are usually much shorter than the detailed captions. Thus, in this task, we do not evaluate Relation Maps Attention and sequence-level training with RL.

Table~\ref{tab:res1} shows the performances of different models for generating high-level captions. 
It is observed that Label Maps Attention can effectively improve the model performances under different metrics. This observation indicates that, different from natural image captioning, features specific to figures, such as labels, can be utilized to boost the model's performance. 


\subsection{Results of Generating Detailed Captions}
We further evaluate the proposed models for the task of generating detailed captions. 
For generating detailed captions, it is important to discover the relations between the labels in the figures, and generate the long sequences of captions. Thus, we further validate the improvements by introducing Relation Maps Attention and sequence-level training by RL.


Table~\ref{tab:res2} shows the performances of different models in generating detailed captions. There are several observations.
First, in this task we observe similar improvements using Label Maps Attention, compared to the results of high-level caption generation in Table~\ref{tab:res1}.
Second, in most cases the performance of CNN-LSTM+Att\_F+Att\_L is better in generating high-level captions, than its performance in generating detailed captions. This indicates that compared to generating high-level captions, generating detailed captions is usually more challenging: in the latter task, we need to model the relations between the labels of figures and handle the long sequence generation.
Third, we achieve significant improvements when introducing Relation Maps Attention and RL. This validates that Relation Maps Attention and RL can effectively model the relations between the labels of figures and the long sequence generation, in the task of generating detailed captions.

\subsection{Discussions}

Experimental results show that the proposed Attention Models for figure captioning are capable of improving the quality of generated captions.
Compared with the baseline model CNN-LSTM, we observe that models that use Attention Models achieve better performance on both FigCAP-H and FigCAP-D. This result indicates that attention-based models are useful for figure captioning.
Second, we found that the effects of \textit{Att\_F} is more higher in FigCAP-D than FigCAP-H. It indicates that generating high-level descriptions does not actually need complex Attention Models since it is more likely a classification task which can be accomplished based on general information of the figure.
In addition, we find that Relation Maps $\bm{R}$ are useful if descriptions about relations of a figure's labels are desired (\emph{e.g.}, Bar A is higher than Bar B; Bar C has the largest value).
Furthermore, with RL we can alleviate the exposure bias issue and directly optimize the evaluation metric used at the inference time.
This enables us to achieve better performance in the generation of long captions.





\section{Conclusion}
In this work, we investigated the problem of figure captioning. 
First, we presented a new dataset, FigCAP, for this figure captioning task, based on FigureQA. Second, we propose two novel attention mechanisms. To achieve accurate generation of labels in figures, we propose Label Maps Attention. To discover the relations between figure labels, we propose Relation Maps Attention. Third, to handle long sequence generation and alleviate the issue of exposure bias, we utilize sequence-level training with reinforcement learning. Experimental results show that the proposed models can effectively generate captions for figures under several metrics.

\bibliography{mybib}
\bibliographystyle{acl_natbib}

\end{document}